\pdfoutput=1

\documentclass[11pt]{article}

\usepackage{acl}

\usepackage{times}
\usepackage{latexsym}
\usepackage{amsmath}
\usepackage{booktabs}
\usepackage{float}
\usepackage[T1]{fontenc}

\usepackage[utf8]{inputenc}

\usepackage{microtype}
\usepackage{hyperref}

\usepackage{inconsolata}

\usepackage{graphicx}

\title{University of 
Indonesia at SemEval-2025 Task 11: Evaluating State-of-the-Art Encoders for Multi-Label Emotion Detection}

\begin{document}
\maketitle
\begin{abstract}
This paper presents our approach for SemEval 2025 Task 11 Track A, focusing on multilabel emotion classification across 28 languages. We explore two main strategies: fully fine-tuning transformer models and classifier-only training, evaluating different settings such as fine-tuning strategies, model architectures, loss functions, encoders, and classifiers. Our findings suggest that training a classifier on top of prompt-based encoders such as mE5 and BGE yields significantly better results than fully fine-tuning XLMR and mBERT. Our best-performing model on the final leaderboard is an ensemble combining multiple BGE models, where CatBoost serves as the classifier, with different configurations. This ensemble achieves an average F1-macro score of 56.58 across all languages.
\end{abstract}

\section{Introduction}
This paper presents the University of Indonesia's multi-label emotion classification system for all 28 languages included in SemEval 2025 Task 11 Track A \cite{semeval2025}. The task focuses on recognizing multiple emotions expressed in text across diverse linguistic and cultural contexts.

Language is a rich and complex medium for conveying emotions \cite{emotion1,emotion2}. However, emotional expression and interpretation vary widely across individuals, even within the same cultural or social background. This variability introduces inherent uncertainty in accurately inferring emotions from textual cues.

Emotion recognition is a challenging task that involves multiple subproblems, such as identifying the speaker’s emotional state, detecting emotions embedded in text, and analyzing the emotional impact on readers \cite{emotion3,emotion4}. Addressing these challenges requires models that can handle multiple emotional labels accurately.

To address this problem, we explore both classifier-only training and end-to-end fine-tuning strategies. Our approach leverages state-of-the-art encoder-based architectures, including Jina, BGE, and multilingual-E5 (mE5) \cite{jina,bge,me5}. These models are pretrained to generate high-quality embeddings, improving classification performance. We experiment with both pre-trained embeddings combined with machine learning classifiers and fine-tuning transformer-based models with specialized loss functions such as Focal Loss and Asymmetric Loss to mitigate class imbalance \cite{asymetric, focalloss}.

Our key findings indicate that embedding-based methods with tree-based classifiers, where we freeze the classifier, particularly BGE combined with CatBoost, outperform fine-tuning approaches for multi-label emotion classification. Specifically, employing separate prompts for each emotion in BGE leads to a improvement in F1-Macro scores. Finally, ensembling enhances the model's robustness, as reflected in our final submission, which shows an improvement compared to using a single model.




\section{Related Works}
This task focuses on multilingual multilabel emotion classification using the BRIGHTER dataset \cite{BRIGHTER}, which includes predominantly low-resource languages from Africa, Asia, Eastern Europe, and Latin America. These instances, annotated by fluent speakers, span multiple domains, presenting unique challenges due to both multilinguality and the complexity of multilabel classification.

Recent advancements in decoder-based models such as LLaMA, GPT, DeepSeek, and Qwen \cite{gpt, gpt4, deepseek, qwen2, llama3}, alongside the widespread use of the BERT family of models \cite{bert, roberta, xlmr}, have demonstrated strong performance in multilingual natural language processing (NLP) tasks. Prior research \cite{evaluatingmultilabel, BRIGHTER} has leveraged these architectures for emotion classification, yet the exploration of advanced encoder-based models like Jina, BGE, and mE5 \cite{jina, bge, me5} remains limited. These models have performed exceptionally well in embedding benchmarks such as MTEB \cite{mteb}, suggesting their potential for our task.

Multilabel classification poses distinct methodological challenges. A traditional approach is Binary Relevance (BR), where separate models are trained for each label \cite{BR}. More recent strategies leverage BERT-based architectures to enable multi-output classification, predicting multiple labels simultaneously \cite{biomultilabel}. Another technique incorporates the [SEP] token to convert multilabel classification into a sequence-labeling task, effectively treating it as a single-label problem \cite{sep}.

A persistent challenge in multilabel classification is class imbalance \cite{imbalancemultilabel}. Unlike standard classification tasks, conventional stratification techniques do not naturally extend to multilabel settings. Iterative stratification methods \cite{stratify} offer a partial solution, while alternative techniques such as weighted loss functions \cite{weighted}, focal loss, and asymmetric loss \cite{focalloss, asymetric} help mitigate imbalance in deep learning models.

Linguistic diversity further complicates multilingual emotion classification. Given the authors' limited language proficiency, exhaustive linguistic analysis across all dataset languages is infeasible. To address this, we explore two approaches, training models separately for each language or collectively across all languages, following prior work \cite{PEAR}.

Our work builds on these foundations by investigating the underexplored potential of advanced encoder-based models in multilingual multilabel emotion classification. By combining these models with effective imbalance-handling techniques and leveraging external linguistic resources, we aim to advance the state of multilingual emotion classification beyond existing methodologies.

\section{System Overview}
\begin{figure*}[t]
  \centering
  \makebox[\textwidth]{\includegraphics[width=1.2\textwidth]{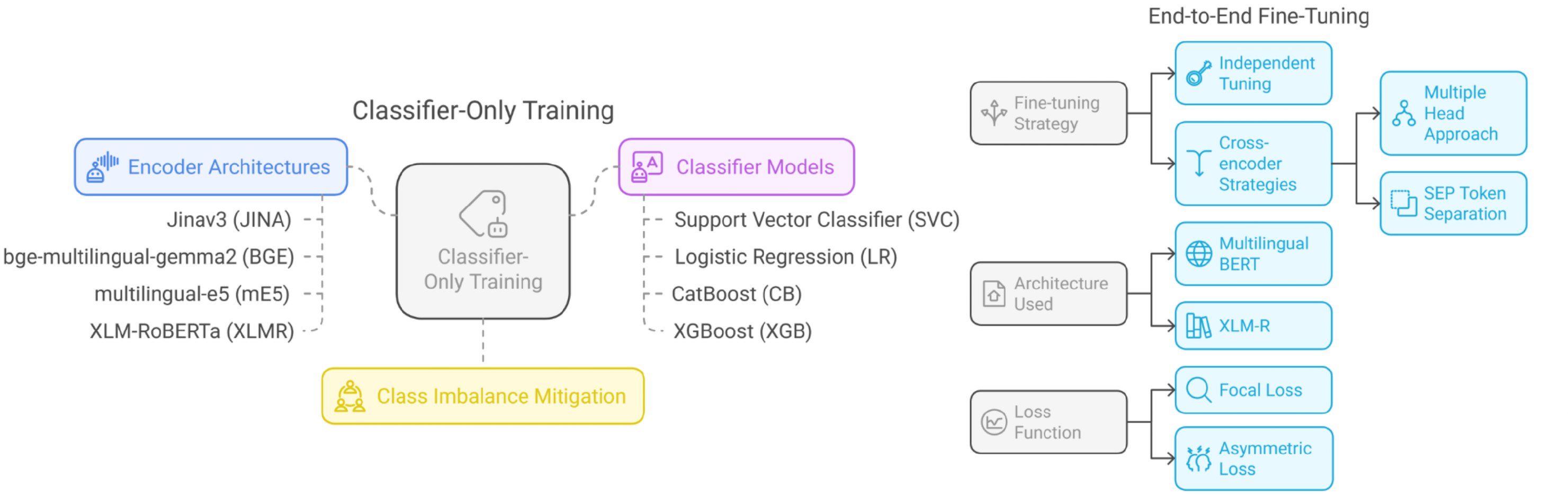}}
  \caption{Our system overview}
  \label{fig:experiments}
\end{figure*}

\subsection{Classifier-Only Training} 
In this approach, we leverage and freeze pre-trained encoders to extract feature representations from text and train classifiers separately for emotion prediction.

\textbf{Utilized Encoder Architectures.} The encoders used in our experiments include \texttt{Jinav3 (JINA)}, \texttt{bge-multilingual-gemma2 (BGE)}, \texttt{multilingual-e5 (mE5)}, and \texttt{XLM-RoBERTa (XLMR)} \cite{jina,bge,me5,xlmr}.

\textbf{Classifier Models.} We explore multiple machine learning models, including \texttt{Support Vector Classifier (SVC)}, \texttt{Logistic Regression(LR)}, \texttt{CatBoost(CB)}, and \texttt{XGBoost(XGB)} as classification models \cite{xgboost, svm, catboost}. To mitigate the imbalance in emotion categories, we employ class weighting to improve the representation of minority classes during training, as defined by the following formula.

\begin{equation}
    w_i = \frac{N}{|C_i| \times k}
\label{weight forumula}
\end{equation}

Where:
\( w_i \) is the weight for class \( i \),
\( N \) is the total number of samples, 
\( |C_i| \) is the number of samples in class \( i \),
\( k \) is the total number of classes.
\subsection{End-to-End Fine-Tuning}

\textbf{Fine-tuning Strategy.} The first type of model involves fine-tuning independently for each emotion category (BR). For the cross-encoder model, we explore two strategies:

\begin{enumerate}  
    \item \textbf{Multiple Head Approach.} A single output layer predicts all emotion categories simultaneously. The model outputs independent probabilities for each emotion using a sigmoid activation function:
    \begin{equation}
    p(y_i | x) = \sigma(W_i x + b_i)
    \end{equation}
    where \( W_i \) and \( b_i \) are the weights and biases for emotion \( i \), and \(\sigma\) is the sigmoid activation function. This configuration is referred to as MultipleOutput \texttt{(MO)}.
    
    \item \textbf{[SEP] Token Separation.} Each input is formatted as \texttt{<sentence> [SEP] <emotion>}, treating the problem as a binary classification for each emotion. This forces the model to consider the relationship between the sentence and a specific emotion. This configuration is referred to as \texttt{SEP}.
\end{enumerate}

\textbf{Architecture Used.} In this experiment, we employ multilingual BERT \texttt{(mBERT)} and \texttt{XLMR} as the underlying architectures for fine-tuning. These models serve as the backbone for our emotion classification framework, leveraging their multilingual pretraining to enhance contextual understanding across diverse languages.

\textbf{Loss Function.} Due to the imbalance of the data set, we employ Focal Loss and Asymmetric Loss:
\begin{enumerate}
    \item \textbf{Focal Loss (FL).} Focal Loss is designed to focus on difficult examples by down-weighting well-classified ones \cite{focalloss}. The formula is:
    \begin{equation}
    FL(p_t) = -\alpha (1 - p_t)^\gamma \log p_t
    \end{equation}
    where \( p_t \) is the predicted probability for the correct class, and \( \alpha \) and \( \gamma \) are parameters controlling class imbalance and focusing strength. Here, we set \( \alpha \) based on the class weight (as in Formula \ref{weight forumula}), and \( \gamma = 2 \).
    
    \item \textbf{Asymmetric Loss (AL).} Asymmetric Loss by applying different focusing strengths for positive and negative samples \cite{asymetric}. The formula is:
    \begin{equation}
    \begin{cases}
    L^+ = (1 - p)^{\gamma^+} \log(p) \\
    L^- = p^{\gamma^-} \log(1 - p)
    \end{cases}
    \end{equation}
    where \( \gamma^+ \) and \( \gamma^- \) control the focusing for positive and negative examples, respectively. For this task, we set \( \gamma^+ = 0 \) and \( \gamma^- = 4 \) as per the original paper. Additionally, to account for shifted probabilities, we use a margin \( m \) such that the probability \( p_m \) is:
    \begin{equation}
    p_m = \max(p - m, 0)
    \end{equation}
    where \( m = 0.05 \). The negative loss term is then adjusted as:
    \begin{equation}
    L^- = (p_m)^{\gamma^-} \log(1 - p_m)
    \end{equation}
\end{enumerate}

\section{Experiment Setting} 

\paragraph{Language \& Data Splits.} We utilize both multilingual and monolingual settings. In the multilingual setting, all available languages are incorporated during training \texttt{All}, while in the monolingual setting, only the target language is used \texttt{LANG}. We split the data into training and validation sets in an 80:20 ratio using iterative stratification \cite{imbalancemultilabel} to ensure an equal distribution of labels. 


\paragraph{Computational Power Used.} We use different machines for different experiments. Lightweight experiments, such as running tree-based models, are conducted using Kaggle’s free GPU, while heavier tasks, such as inferencing with \texttt{BGE, mE5, JINA}, are performed on an RTX 4090 rented from the Vast.ai platform.

\paragraph{Hyperparameter Settings.} In both approaches, no additional hyperparameter tuning is performed, ensuring that all models share a consistent set of parameters across experiments. The details are provided in Appendix Table~\ref{tab:hyperparams}. 

\paragraph{Encoder Settings.} \texttt{XLMR} is direct use require no additional settings. \texttt{JINA} required to set task and prompt\_name parameter which both are set to 'classification'. \texttt{mE5} and \texttt{BGE} require prompt which we adapt from the original papers. Specifically, \texttt{mE5} and \texttt{BGE} (\texttt{V1}), we used general prompts asking to detect multiple emotions at once. Based on ablation studies, we hypothesized that specifying a single emotion per prompt could improve performance. This led to \texttt{BGE} (\texttt{V2}), where each query focuses on one emotion. Results suggest that targeted instructions better guide the model's representation. Prompt can be seen in \hyperref[tab:prompts]{Appendix} section. 


\section{Result}

\subsection{Development}
In this section, we analyze the average F1 Macro scores across all languages to guide our model selection and evaluation based on our results on the development set. Evaluation table for \hyperref[tab:model_scores]{classifier-only} and \hyperref[tab:additional_model_scores]{others} are on the Appendix section.

\paragraph{Quantitative Evaluation using Hypothesis Testing.} We employ non-parametric tests to assess whether significant differences exist between model configurations. For paired scenarios, such as \texttt{FL} vs. \texttt{AL} and \texttt{ALL} vs. \texttt{LANG}, we use the Wilcoxon signed-rank test, while for unpaired scenarios, we apply the Mann-Whitney U test \cite{mann, wiloxon}. In paired comparisons, we ensure that only the relevant factor varies while keeping the architecture consistent.

\paragraph{\texttt{FL} vs \texttt{AL}.} The Wilcoxon signed-rank test yielded no significant difference (\( W = 4, p = 0.875 \)), suggesting that both loss functions perform similarly in addressing class imbalance within multi-label emotion detection. Despite their theoretical differences, our results show that neither approach provides a clear advantage. This finding underscores the importance of considering other factors, such as model architecture, in performance optimization.

\paragraph{\texttt{ALL} language vs \texttt{LANG}.} The Wilcoxon signed-rank test showed no significant difference (\( W = 48, p = 0.06 \)) between training on all languages (\texttt{ALL}) and training on a specific language (\texttt{LANG}). This suggests that multilingual training does not necessarily improve performance compared to language-specific models for this task. Moreover, training on \texttt{LANG} is computationally more efficient, as it operates on a smaller, more targeted dataset, making it a practical choice in resource-constrained settings. Additionally, the results suggest that the model’s ability to leverage cross-language associations, a key advantage of multilingual architectures, does not play a significant role in this task.

\paragraph{LLM Prompt Based Encoder with Classifier Outperform Fully Finetuned Transformer.} 
The Mann--Whitney \(U\) test indicates a significant difference (\(U = 456,\,p < 0.001\)), with prompt-based encoder models (BGE and mE5) outperforming all others. Their average F1 Macro scores, \(47.3\%\) for BGE and \(37.7\%\) for fully fine-tuned models, reveal a clear gap. This stems from BGE and mE5’s demonstrated superiority on the MMTEB \cite{MMTEB} multilingual embedding benchmark, which attests to their stronger multilingual representations; fine-tuning on low-resource task data cannot match this pre-validated embedding quality.

\paragraph{BGE as the Overall Best Result.} The statistical test yielded a significant result (\( W = 205, p = 0.009 \)), confirming that BGE-based models significantly outperform non-BGE models, particularly XLMR and mBERT, despite requiring less computational power. These findings reinforce the effectiveness of BGE’s architecture in capturing emotion-related semantics, making it a strong candidate for future research in multilingual emotion classification.



\paragraph{Different prompt lead to different results.} We observe that modifying the prompt from general to slightly more specific consistently improves performance. Although this experiment was conducted only on CB models with two samples, the observed differences are notable, with F1 Macro scores increasing from 5.3\% to 5.5\% and from 54.0\% to 55.0\%. These results suggest that refining prompts can enhance model effectiveness.

\subsection{Submission}  

For this shared task, we have two types of submissions:  

\begin{itemize}
    \item \textbf{Model V1}: The highest score model, \texttt{BGEV2-CB-ALL}.  
    \item \textbf{Model V2}: An ensemble of four models: \texttt{BGEV2-CB-ALL}, \texttt{BGE-CB-LANG}, \texttt{BGE-CB-LANG}, \texttt{BGE-CB-ALL}.  
\end{itemize}

\begin{table}[H]
    \centering
    \renewcommand{\arraystretch}{1.2}
    \setlength{\tabcolsep}{8pt} 
    \begin{tabular}{lccc}
        \hline
        \textbf{Lang} & \textbf{Model V1} & \textbf{Model V2} & \textbf{Qwen2.5} \\
        \hline
        afr  & 53.99 & 54.57 & \textbf{\textcolor{orange}{60.18}} \\
        amh  & 50.29 & \textbf{\textcolor{orange}{51.18}} & - \\
        deu  & 64.50 & \textbf{\textcolor{orange}{66.16}} & 59.17 \\
        eng  & 72.47 & \textbf{\textcolor{orange}{74.94}} & 55.72 \\
        esp  & 75.60 & \textbf{\textcolor{orange}{79.53}} & 72.33 \\
        hin  & 79.21 & \textbf{\textcolor{orange}{86.05}} & 79.73 \\
        mar  & \textbf{\textcolor{orange}{84.73}} & 81.60 & 74.58 \\
        orm  & 40.52 & \textbf{\textcolor{orange}{46.25}} & - \\
        ptbr & 55.27 & \textbf{\textcolor{orange}{56.88}} & 51.60 \\
        rus  & 76.29 & \textbf{\textcolor{orange}{84.37}} & 73.08 \\
        som  & 42.79 & \textbf{\textcolor{orange}{43.73}} & - \\
        sun  & 42.86 & \textbf{\textcolor{orange}{43.17}} & 42.67 \\
        tat  & 59.26 & \textbf{\textcolor{orange}{60.39}} & 51.58 \\
        tir  & 37.55 & \textbf{\textcolor{orange}{40.03}} & - \\
        arq  & 52.70 & \textbf{\textcolor{orange}{54.99}} & 37.78 \\
        ary  & 51.99 & \textbf{\textcolor{orange}{53.50}} & 52.76 \\
        chn  & 61.71 & \textbf{\textcolor{orange}{62.87}} & 55.23 \\
        hau  & 59.98 & \textbf{\textcolor{orange}{64.43}} & 43.79 \\
        kin  & 43.34 & \textbf{\textcolor{orange}{48.35}} & 31.96 \\
        pcm  & 58.35 & \textbf{\textcolor{orange}{60.45}} & 38.66 \\
        ptmz & 38.24 & \textbf{\textcolor{orange}{42.72}} & 40.44 \\
        swa  & \textbf{\textcolor{orange}{37.88}} & 37.55 & 27.36 \\
        swe  & 56.72 & \textbf{\textcolor{orange}{57.84}} & 48.89 \\
        ukr  & 54.99 & \textbf{\textcolor{orange}{63.36}} & 54.76 \\
        vmw  & 10.74 & 13.55 & \textbf{\textcolor{orange}{20.41}} \\
        yor  & 26.60 & \textbf{\textcolor{orange}{29.05}} & 24.99 \\
        ibo  & 47.93 & \textbf{\textcolor{orange}{49.58}} & 37.40 \\
        ron  & \textbf{\textcolor{orange}{74.78}} & 73.80 & 68.18 \\
        \hline
    \end{tabular}
    \caption{\small Test set comparison of our models with the Qwen2.5-72B decoder model, which has the highest average F1 Macro score in the BRIGHTER paper \cite{BRIGHTER}.}
    \label{tab:model_comparison}
\end{table}

\vspace{-5mm}

\begin{equation}
    v_i =
    \begin{cases} 
    1, & \text{if } y_i = 1 \\ 
    -1, & \text{if } y_i = 0
    \end{cases}
\end{equation}

The final predicted label is then given by:

\begin{equation}
    \hat{y} =
    \begin{cases} 
    1, & \text{if } s > 0 \\ 
    0, & \text{otherwise}
    \end{cases}
\end{equation}

We use weighted voting for ensemble predictions, assigning weights based on development set performance and handling zero weights, with the final prediction determined as follows.

\begin{equation}
    s = \sum_{i=1}^{N} w_i \cdot v_i
\end{equation}

where \( s \) is the aggregated weighted score, \( N \) is the number of models, \( w_i \) is the weight of the \( i \)-th model based on its development set score, and \( v_i \) is the adjusted prediction:

This ensures that zero predictions contribute negatively instead of being ignored, and the final decision is based on the sign of the weighted sum.

Based on Table \ref{tab:model_comparison}, Model V2, an ensemble, outperforms Model V1 in 25 out of 28 languages. Using the Wilcoxon signed-rank test, we obtain \( W = 285.0 \) and \( p < 0.001 \), indicating a statistically significant improvement over Qwen2.5-72B \cite{BRIGHTER}.

\section{Limitations}
The limitation of our study is the lack of extensive qualitative analysis due to limited language proficiency. Since we do not fully understand many of the languages in the dataset, our analysis primarily relies on quantitative methods.

\section{Conclusion}
Our study demonstrates that classifier-based approaches with prompt-based encoders, particularly BGE and multilingual-E5 (mE5), outperform fully fine-tuned transformer models for multilingual multi-label emotion classification. Our best-performing model, BGE with CatBoost and emotion-specific prompting, achieved the highest average F1-Macro scores across languages in our experiment. Additionally, an ensemble of multiple BGE-based models further improved performance, significantly surpassing the best decoder-based model from prior work. These results highlight the strength of high-quality embeddings combined with tree-based classifiers for emotion detection tasks.  

Our findings also show that multilingual training does not provide a clear advantage over monolingual models. Furthermore, minor prompt modifications led to measurable gains, emphasizing the importance of prompt engineering. Overall, our study suggests that leveraging strong embedding models with efficient classifiers is a more effective strategy than full transformer fine-tuning for multi-label emotion classification across diverse languages.

\section{Appendix}
\label{sec:appendix}

\begin{table*}[h]
    \centering
    \begin{tabular}{l l}
        \hline
        \textbf{Model Name} & \textbf{Prompt} \\  
        \hline
        {mE5} & \texttt{Instruct: Classify the emotions expressed in the given text snippet} \\
        & \texttt{by identifying whether each of the following emotions is present:} \\
        & \texttt{joy, sadness, anger, surprise, and disgust.} \\
        \\
        & \texttt{Query: \{\{INPUT\}\}} \\
        \\
        {BGEV1} & \texttt{<instruct> Represent this text for identifying the presence of emotions:} \\
        & \texttt{joy, sadness, anger, surprise, and disgust} \\
        \\
        & \texttt{<query> \{\{INPUT\}\}} \\
        \\
        {BGEV2} & \texttt{<instruct> Represent this text for identifying the presence of the} \\
        & \texttt{emotion \{\{EMOTION\}\}} \\ 
        \\
        & \texttt{<query> \{\{INPUT\}\}} \\
        \hline
    \end{tabular}
    \caption{Prompt formulations used for mE5 and BGE models}
    \label{tab:prompts}
\end{table*}

\begin{table*}[h]
    \centering
    \begin{tabular}{l l}
        \hline
        \textbf{Model} & \textbf{Hyperparameter} \\  
        \hline
        {mBERT, XLMR} & Learning Rate: $3 \times 10^{-5}$ \\
                                      & Training Batch Size: 32 \\
                                      & Evaluation Batch Size: 8 \\
                                      & Seed: 42 \\
                                      & LR Scheduler Type: Linear \\
                                      & LR Scheduler Warmup Steps: $0.1 \times$ total train steps \\
                                      & Number of Epochs: 4 \\
        \hline
    \end{tabular}
    \caption{Hyperparameter settings for mBERT and XLM-R models.}
    \label{tab:hyperparams}
\end{table*}

\begin{table*}[h]
    \scriptsize
    \centering
    \begin{tabular}{lcccccc}
        \hline
        \textbf{language} & \textbf{BGEV1-CB-ALL} & \textbf{BGEV2-CB-ALL} & \textbf{BGEV1-CB-LANG} & \textbf{BGEV2-CB-LANG} & \textbf{BGEV1-LR-ALL} & \textbf{BGEV1-LR-LANG} \\ \hline
        afr & 49.64 & 62.10 & 56.41 & 53.40 & 48.07 & 52.02 \\
        amh & 47.72 & 50.70 & 47.22 & 48.89 & 50.66 & 53.18 \\
        arq & 54.03 & 56.56 & 57.40 & 59.83 & 51.79 & 57.09 \\
        ary & 50.92 & 50.02 & 46.54 & 50.02 & 43.08 & 51.02 \\
        chn & 60.24 & 60.66 & 59.79 & 59.76 & 56.02 & 61.59 \\
        deu & 62.11 & 66.05 & 58.42 & 63.66 & 62.10 & 60.35 \\
        eng & 72.36 & 71.75 & 76.36 & 74.41 & 64.88 & 74.29 \\
        esp & 76.30 & 80.14 & 78.40 & 82.60 & 73.83 & 76.79 \\
        hau & 59.21 & 58.28 & 65.54 & 63.93 & 57.73 & 64.25 \\
        hin & 76.83 & 80.66 & 81.62 & 85.81 & 76.33 & 83.96 \\
        ibo & 47.15 & 45.15 & 45.93 & 43.34 & 44.67 & 45.22 \\
        kin & 47.95 & 49.05 & 47.75 & 47.08 & 40.94 & 41.69 \\
        mar & 90.03 & 82.53 & 92.15 & 84.74 & 89.81 & 91.33 \\
        orm & 36.28 & 40.97 & 41.15 & 41.02 & 43.77 & 42.65 \\
        pcm & 57.01 & 57.09 & 56.01 & 58.13 & 54.26 & 49.97 \\
        ptbr & 53.53 & 54.56 & 51.94 & 54.30 & 53.04 & 49.98 \\
        ptmz & 47.97 & 43.12 & 43.37 & 38.39 & 43.52 & 44.32 \\
        ron & 72.93 & 80.30 & 72.20 & 94.17 & 72.15 & 70.84 \\
        rus & 76.85 & 76.24 & 82.32 & 82.08 & 80.93 & 84.58 \\
        som & 41.51 & 41.76 & 39.76 & 42.14 & 38.53 & 38.95 \\
        sun & 47.75 & 49.87 & 44.28 & 40.36 & 45.46 & 46.17 \\
        swa & 39.07 & 39.10 & 36.20 & 36.22 & 34.61 & 30.21 \\
        swe & 49.31 & 57.20 & 47.48 & 48.98 & 47.22 & 47.93 \\
        tat & 48.72 & 56.24 & 56.94 & 52.84 & 50.11 & 61.59 \\
        tir & 35.04 & 39.40 & 38.70 & 36.62 & 38.38 & 38.49 \\
        ukr & 51.79 & 52.45 & 48.69 & 56.09 & 52.18 & 45.43 \\
        vmw & 14.95 & 16.47 & 17.98 & 19.58 & 16.74 & 18.79 \\
        yor & 31.48 & 32.68 & 26.74 & 32.65 & 33.76 & 29.42 \\
        \hline
        \textbf{average} & 53.52 & 55.40 & 54.19 & 55.39 & 52.31 & 54.00 \\
        \hline
    \end{tabular}
    \caption{Detailed performance comparison across models on development data – Part 1.}
    \label{tab:model_comparison}
\end{table*}

\begin{table*}[h]
    \centering
    \renewcommand{\arraystretch}{1.2}
    \begin{tabular}{p{5cm} c}
        \toprule
        \textbf{Model} & \textbf{F1 Macro (\%)} \\
        \midrule
        BGEV1-CB-ALL & 53.52 \\
        \textbf{\textcolor{orange}{BGEV2-CB-ALL}} & \textbf{\textcolor{orange}{55.40}} \\
        BGEV1-CB-LANG & 54.19 \\
        BGEV2-CB-LANG & 55.39 \\
        BGEV1-LR-ALL & 52.31 \\
        BGEV1-LR-LANG & 54.00 \\
        BGEV1-SVC-ALL & 18.13 \\
        BGEV1-SVC-LANG & 22.38 \\
        BGEV1-XGB-ALL & 48.41 \\
        BGEV1-XGB-LANG & 48.32 \\
        mE5-CB-ALL & 52.20 \\
        mE5-CB-LANG & 52.49 \\
        mE5-LR-ALL & 49.71 \\
        mE5-LR-LANG & 49.63 \\
        mE5-SGB-LANG & 47.46 \\
        mE5-SVC-ALL & 41.05 \\
        mE5-SVC-LANG & 42.42 \\
        mE5-XGB-ALL & 47.97 \\
        JINA-CB-ALL & 44.78 \\
        JINA-CB-LANG & 46.32 \\
        JINA-LR-ALL & 43.16 \\
        JINA-LR-LANG & 49.05 \\
        JINA-SVC-ALL & 35.40 \\
        JINA-SVC-LANG & 40.08 \\
        JINA-XGB-ALL & 36.38 \\
        JINA-XGB-LANG & 38.54 \\
        XLMR-CB-ALL & 38.48 \\
        XLMR-CB-LANG & 38.38 \\
        XLMR-LR-ALL & 40.52 \\
        XLMR-SVC-ALL & 25.47 \\
        XLMR-SVC-LANG & 33.96 \\
        XLMR-LR-LANG & 46.99 \\
        XLMR-XGB-ALL & 30.88 \\
        XLMR-XGB-LANG & 29.30 \\
        \bottomrule
    \end{tabular}
    \caption{Performance scores of the classifier-only training model on the test set}
    \label{tab:model_scores}
\end{table*}

\begin{table*}[h]
    \centering
    \renewcommand{\arraystretch}{1.2}
    \begin{tabular}{p{5cm} c}
        \toprule
        \textbf{Model} & \textbf{F1 Macro (\%)} \\
        \midrule
        mBERT-BR-LANG-FL & 46.71 \\
        \textbf{\textcolor{orange}{mBERT-MO-ALL-AL}} & \textbf{\textcolor{orange}{47.10}} \\
        mBERT-MO-ALL-FL & 39.95 \\
        mBERT-MO-LANG-AL & 42.39 \\
        mBERT-MO-LANG-FL & 40.13 \\
        mBERT-SEP-LANG & 39.54 \\
        XLMR-BR-LANG-FL & 45.61 \\
        XLMR-MO-ALL-AL & 21.74 \\
        XLMR-MO-ALL-FL & 42.38 \\
        XLMR-MO-LANG-AL & 27.85 \\
        XLMR-MO-LANG-FL & 25.61 \\
        XLMR-SEP-LANG-FL & 21.25 \\
        \bottomrule
    \end{tabular}
    \caption{Performance score of the fully fine-tuned model on the development set}
    \label{tab:additional_model_scores}
\end{table*}

\begin{table*}[h]
    \centering
    \scriptsize
    \begin{tabular}{lcccccc}
        \hline
        \textbf{language} & \textbf{BGE-SVM-ALL} & \textbf{BGE-SVM-LANG} & \textbf{BGE-XGB-ALL} & \textbf{BGE-XGB-LANG} & \textbf{mE5-CB-ALL} & \textbf{mE5-CB-LANG} \\ \hline
        afr & 11.35 & 23.52 & 39.24 & 50.09 & 50.93 & 53.06 \\
        amh & 22.73 & 26.55 & 41.27 & 39.04 & 54.63 & 54.17 \\
        arq & 29.81 & 40.68 & 47.79 & 54.78 & 49.21 & 52.27 \\
        ary & 17.02 & 21.17 & 43.91 & 42.36 & 44.16 & 48.03 \\
        chn & 20.75 & 20.63 & 58.84 & 53.82 & 56.61 & 53.38 \\
        deu & 24.23 & 26.19 & 55.61 & 57.51 & 56.15 & 54.74 \\
        eng & 27.47 & 39.33 & 60.89 & 75.70 & 75.25 & 75.16 \\
        esp & 22.86 & 24.04 & 77.48 & 77.16 & 73.85 & 76.81 \\
        hau & 21.10 & 26.35 & 59.34 & 62.61 & 53.91 & 55.93 \\
        hin & 15.18 & 19.10 & 83.52 & 83.62 & 70.69 & 74.44 \\
        ibo & 18.08 & 21.65 & 41.13 & 42.19 & 40.85 & 40.68 \\
        kin & 10.60 & 17.58 & 40.01 & 40.72 & 44.50 & 45.59 \\
        mar & 17.10 & 26.35 & 92.29 & 90.94 & 88.72 & 91.02 \\
        orm & 15.78 & 20.56 & 32.58 & 33.19 & 40.61 & 39.92 \\
        pcm & 24.03 & 29.72 & 54.47 & 48.88 & 50.37 & 50.21 \\
        ptbr & 18.34 & 21.29 & 48.76 & 36.60 & 48.97 & 49.21 \\
        ptmz & 13.75 & 14.13 & 42.70 & 40.75 & 47.76 & 45.86 \\
        ron & 28.27 & 34.57 & 68.62 & 68.75 & 69.47 & 72.53 \\
        rus & 18.13 & 19.00 & 82.70 & 81.57 & 79.85 & 80.36 \\
        som & 12.06 & 18.74 & 27.44 & 31.93 & 38.83 & 38.00 \\
        sun & 15.34 & 24.76 & 34.29 & 35.89 & 43.94 & 41.75 \\
        swa & 11.92 & 16.12 & 28.64 & 21.62 & 28.25 & 26.30 \\
        swe & 16.52 & 18.63 & 46.69 & 40.98 & 54.45 & 52.97 \\
        tat & 18.55 & 19.65 & 46.71 & 49.92 & 66.35 & 61.82 \\
        tir & 19.52 & 15.98 & 29.35 & 26.78 & 43.63 & 42.31 \\
        ukr & 11.69 & 15.27 & 49.88 & 47.71 & 53.71 & 50.04 \\
        vmw & 15.21 & 13.75 & 3.62 & 1.96 & 8.71 & 18.27 \\
        yor & 10.27 & 11.40 & 17.64 & 15.81 & 27.12 & 24.93 \\
        \hline
        \textbf{average} & 18.13 & 22.38 & 48.41 & 48.32 & 52.20 & 52.49 \\
        \hline
    \end{tabular}
    \caption{Detailed performance comparison across models on development data – Part 2.}
    \label{tab:model_comparison}
\end{table*}

\begin{table*}[h]
    \centering
    \scriptsize
    \begin{tabular}{lcccccc}
        \hline
        \textbf{language} & \textbf{mE5-LR-ALL} & \textbf{mE5-LR-LANG} & \textbf{mE5-SGB-LANG} & \textbf{mE5-SVM-ALL} & \textbf{mE5-SVM-LANG} & \textbf{mE5-XGB-ALL} \\ \hline
        afr & 49.04 & 47.42 & 38.07 & 47.89 & 49.81 & 48.05 \\
        amh & 55.58 & 52.61 & 46.31 & 40.34 & 45.24 & 47.68 \\
        arq & 52.56 & 51.41 & 47.41 & 38.47 & 38.63 & 42.89 \\
        ary & 38.16 & 40.48 & 42.80 & 39.30 & 27.74 & 42.30 \\
        chn & 55.83 & 52.69 & 49.34 & 46.64 & 45.73 & 53.26 \\
        deu & 53.52 & 56.84 & 56.76 & 47.09 & 51.66 & 58.14 \\
        eng & 70.50 & 71.08 & 73.95 & 66.80 & 66.54 & 66.64 \\
        esp & 70.85 & 74.80 & 76.83 & 60.54 & 71.80 & 75.12 \\
        hau & 47.55 & 53.34 & 53.30 & 32.00 & 48.55 & 51.22 \\
        hin & 61.89 & 66.07 & 79.76 & 53.35 & 58.46 & 77.97 \\
        ibo & 35.98 & 37.96 & 38.06 & 25.76 & 36.35 & 36.01 \\
        kin & 33.05 & 39.97 & 43.33 & 26.90 & 33.98 & 37.48 \\
        mar & 80.66 & 78.42 & 90.88 & 69.40 & 87.23 & 91.03 \\
        orm & 37.33 & 39.22 & 35.88 & 28.15 & 28.82 & 32.08 \\
        pcm & 48.62 & 50.18 & 42.79 & 44.83 & 40.96 & 46.06 \\
        ptbr & 52.69 & 47.42 & 41.04 & 48.09 & 39.98 & 38.04 \\
        ptmz & 41.45 & 37.24 & 38.96 & 29.93 & 28.84 & 46.19 \\
        ron & 68.09 & 69.71 & 72.17 & 67.40 & 55.76 & 71.86 \\
        rus & 75.03 & 72.25 & 80.72 & 61.39 & 74.09 & 81.90 \\
        som & 39.38 & 37.86 & 30.08 & 28.65 & 27.77 & 31.71 \\
        sun & 47.91 & 44.91 & 36.95 & 37.73 & 40.53 & 36.49 \\
        swa & 29.84 & 28.20 & 13.71 & 22.50 & 19.01 & 17.58 \\
        swe & 47.60 & 47.50 & 48.02 & 42.99 & 45.07 & 51.27 \\
        tat & 60.46 & 57.84 & 57.67 & 42.76 & 39.52 & 59.10 \\
        tir & 45.19 & 41.59 & 34.04 & 31.39 & 26.86 & 36.93 \\
        ukr & 54.37 & 45.75 & 44.04 & 42.63 & 36.89 & 46.85 \\
        vmw & 11.39 & 20.17 & 00.95 & 05.23 & 02.06 & 01.65 \\
        yor & 27.45 & 26.60 & 15.12 & 21.23 & 19.82 & 17.62 \\
        \hline
        \textbf{average} & 49.71 & 49.63 & 47.46 & 41.05 & 42.42 & 47.97 \\
        \hline
    \end{tabular}
    \caption{Detailed performance comparison across models on development data – Part 3.}
    \label{tab:model_comparison}
\end{table*}

\begin{table*}[h]
    \centering
    \scriptsize
    \begin{tabular}{lcccccc}
        \hline
        \textbf{language} & \textbf{JINA-CB-ALL} & \textbf{JINA-CB-LANG} & \textbf{JINA-LR-ALL} & \textbf{JINA-LR-LANG} & \textbf{JINA-SVM-ALL} & \textbf{JINA-SVM-LANG} \\ \hline
        afr & 42.48 & 27.99 & 40.25 & 41.43 & 35.03 & 23.27 \\
        amh & 50.24 & 48.19 & 48.14 & 50.96 & 42.78 & 43.00 \\
        arq & 51.98 & 45.56 & 50.18 & 52.91 & 45.03 & 36.39 \\
        ary & 42.68 & 46.64 & 39.39 & 46.45 & 32.88 & 33.68 \\
        chn & 53.93 & 53.24 & 51.10 & 56.27 & 46.18 & 44.82 \\
        deu & 52.94 & 52.62 & 53.71 & 57.44 & 40.44 & 50.09 \\
        eng & 62.92 & 68.95 & 63.38 & 69.41 & 58.22 & 61.98 \\
        esp & 67.07 & 69.26 & 64.14 & 71.83 & 59.52 & 66.42 \\
        hau & 44.60 & 50.65 & 38.07 & 49.19 & 20.72 & 38.77 \\
        hin & 61.95 & 72.82 & 58.95 & 68.48 & 47.25 & 59.73 \\
        ibo & 34.11 & 39.24 & 29.34 & 41.31 & 18.52 & 33.68 \\
        kin & 29.07 & 33.43 & 28.69 & 33.92 & 17.45 & 27.85 \\
        mar & 72.27 & 79.63 & 68.98 & 75.58 & 54.47 & 68.92 \\
        orm & 27.61 & 34.31 & 29.62 & 37.25 & 19.78 & 30.84 \\
        pcm & 48.09 & 45.86 & 46.44 & 50.14 & 41.94 & 37.32 \\
        ptbr & 49.09 & 45.28 & 45.97 & 47.52 & 37.72 & 38.50 \\
        ptmz & 44.53 & 40.26 & 40.89 & 47.74 & 30.63 & 33.63 \\
        ron & 68.11 & 69.22 & 67.75 & 70.29 & 62.95 & 66.79 \\
        rus & 63.77 & 74.63 & 59.95 & 71.75 & 49.01 & 56.86 \\
        som & 26.30 & 28.67 & 27.12 & 32.45 & 20.56 & 29.11 \\
        sun & 41.93 & 45.22 & 38.39 & 44.38 & 34.75 & 37.95 \\
        swa & 26.96 & 24.61 & 29.08 & 29.08 & 23.35 & 18.53 \\
        swe & 45.78 & 49.31 & 45.61 & 50.54 & 42.28 & 44.25 \\
        tat & 39.71 & 33.54 & 36.00 & 47.66 & 27.80 & 37.35 \\
        tir & 35.92 & 36.37 & 37.09 & 39.39 & 31.51 & 30.33 \\
        ukr & 41.35 & 42.75 & 37.03 & 45.29 & 30.43 & 34.85 \\
        vmw & 12.78 & 19.97 & 14.65 & 23.36 & 08.57 & 21.30 \\
        yor & 15.53 & 18.72 & 18.67 & 21.34 & 11.43 & 15.92 \\
        \hline
        \textbf{average} & 44.78 & 46.32 & 43.16 & 49.05 & 35.40 & 40.08 \\
        \hline
    \end{tabular}
    \caption{Detailed performance comparison across models on development data – Part 4}
    \label{tab:model_comparison}
\end{table*}

\begin{table*}[h]
    \centering
    \scriptsize
    \begin{tabular}{lcccccc}
        \hline
        \textbf{language} & \textbf{JINA-XGB-ALL} & \textbf{JINA-XGB-LANG} & \textbf{MBERT-BR-LANG} & \textbf{MBERT-MO-ALL-AL} & \textbf{MBERT-MULTIOUT-ALL-FL} & \textbf{MBERT-MO-LANG-AL} \\ \hline
        afr & 32.44 & 22.03 & 36.43 & 44.58 & 36.63 & 40.67 \\
        amh & 44.71 & 42.52 & 24.80 & 28.37 & 30.79 & 27.75 \\
        arq & 37.24 & 45.41 & 47.18 & 48.02 & 44.56 & 45.08 \\
        ary & 35.00 & 33.40 & 34.81 & 38.96 & 33.26 & 37.35 \\
        chn & 48.38 & 50.14 & 53.46 & 57.86 & 45.42 & 53.09 \\
        deu & 43.38 & 44.05 & 46.90 & 53.13 & 46.32 & 40.96 \\
        eng & 52.03 & 66.01 & 62.84 & 63.23 & 51.32 & 60.87 \\
        esp & 69.77 & 71.21 & 69.71 & 66.45 & 55.45 & 61.64 \\
        hau & 31.99 & 38.48 & 61.11 & 52.29 & 38.82 & 49.96 \\
        hin & 72.41 & 69.73 & 60.48 & 66.00 & 48.53 & 60.56 \\
        ibo & 30.41 & 31.82 & 45.44 & 44.71 & 35.57 & 42.18 \\
        kin & 16.44 & 25.07 & 42.32 & 35.42 & 26.31 & 31.89 \\
        mar & 78.28 & 73.08 & 84.35 & 81.86 & 72.11 & 79.82 \\
        orm & 20.04 & 22.11 & 50.79 & 43.49 & 33.36 & 33.69 \\
        pcm & 33.92 & 34.05 & 51.25 & 49.77 & 42.88 & 45.57 \\
        ptbr & 35.09 & 35.13 & 33.71 & 39.52 & 37.05 & 30.56 \\
        ptmz & 34.97 & 35.44 & 41.47 & 41.30 & 31.48 & 37.23 \\
        ron & 55.39 & 71.15 & 65.47 & 72.14 & 66.70 & 67.75 \\
        rus & 69.73 & 70.40 & 73.18 & 75.45 & 59.48 & 70.94 \\
        som & 12.03 & 15.83 & 40.46 & 33.58 & 27.27 & 32.38 \\
        sun & 24.54 & 32.54 & 42.38 & 40.05 & 36.77 & 35.00 \\
        swa & 11.38 & 08.84 & 23.42 & 26.76 & 23.91 & 24.16 \\
        swe & 40.59 & 39.03 & 41.48 & 48.00 & 41.47 & 42.49 \\
        tat & 14.15 & 26.46 & 51.14 & 52.01 & 45.92 & 43.53 \\
        tir & 25.26 & 25.95 & 24.67 & 21.56 & 25.67 & 21.75 \\
        ukr & 40.79 & 33.65 & 41.48 & 51.65 & 41.59 & 33.36 \\
        vmw & 01.62 & 06.07 & 25.41 & 14.63 & 18.07 & 11.87 \\
        yor & 06.67 & 09.52 & 31.70 & 28.03 & 21.82 & 24.96 \\
        \hline
        \textbf{average} & 36.38 & 38.54 & 46.71 & 47.10 & 39.95 & 42.40 \\
        \hline
    \end{tabular}
    \caption{Detailed performance comparison across models on development data – Part 5.}
    \label{tab:model_comparison}
\end{table*}

\begin{table*}[h]
    \centering
    \scriptsize
    \begin{tabular}{lcccccc}
        \hline
        \textbf{language} & \textbf{MBERT-MO-LANG-FL} & \textbf{MBERT-SEP-LANG} & \textbf{XLMR-BR-LANG} & \textbf{XLMR-MO-ALL-AL} & \textbf{XLMR-MO-ALL-FL} & \textbf{XLMR-MO-LANG-AL} \\ \hline
        afr & 41.53 & 30.18 & 42.09 & 21.68 & 45.78 & 06.90 \\
        amh & 30.99 & 23.06 & 27.14 & 35.27 & 50.50 & 30.69 \\
        arq & 48.55 & 42.00 & 44.19 & 24.02 & 48.89 & 38.46 \\
        ary & 34.30 & 39.05 & 34.32 & 22.28 & 37.28 & 24.59 \\
        chn & 41.82 & 48.00 & 44.73 & 32.88 & 52.51 & 29.83 \\
        deu & 48.16 & 46.50 & 48.09 & 31.62 & 51.12 & 32.28 \\
        eng & 57.89 & 62.80 & 71.00 & 27.01 & 57.41 & 45.44 \\
        esp & 58.17 & 64.79 & 74.51 & 29.85 & 57.68 & 39.19 \\
        hau & 46.95 & 50.91 & 52.64 & 21.42 & 44.85 & 34.95 \\
        hin & 48.23 & 49.32 & 76.31 & 23.68 & 52.66 & 21.65 \\
        ibo & 39.29 & 40.35 & 33.57 & 19.80 & 31.28 & 20.49 \\
        kin & 32.45 & 34.92 & 37.70 & 08.30 & 35.17 & 23.49 \\
        mar & 61.75 & 60.71 & 90.56 & 25.50 & 66.14 & 31.96 \\
        orm & 35.12 & 40.27 & 29.60 & 11.49 & 28.84 & 28.33 \\
        pcm & 44.85 & 44.96 & 51.65 & 28.67 & 46.28 & 36.51 \\
        ptbr & 32.59 & 32.98 & 33.60 & 25.55 & 42.96 & 27.91 \\
        ptmz & 27.00 & 24.67 & 41.36 & 16.56 & 32.96 & 13.03 \\
        ron & 69.62 & 63.69 & 72.00 & 38.94 & 69.35 & 46.52 \\
        rus & 57.26 & 69.75 & 79.48 & 27.13 & 56.29 & 31.33 \\
        som & 29.84 & 25.44 & 31.81 & 15.66 & 33.49 & 22.41 \\
        sun & 37.23 & 26.53 & 37.07 & 18.63 & 41.95 & 39.83 \\
        swa & 23.57 & 23.78 & 27.29 & 13.50 & 28.92 & 18.76 \\
        swe & 41.69 & 39.98 & 44.83 & 33.43 & 46.45 & 33.83 \\
        tat & 41.41 & 45.89 & 38.69 & 11.40 & 37.09 & 23.70 \\
        tir & 27.12 & 18.93 & 36.48 & 22.68 & 32.52 & 29.21 \\
        ukr & 25.78 & 36.23 & 46.35 & 15.35 & 35.77 & 18.14 \\
        vmw & 18.39 & 11.20 & 15.82 & 03.41 & 06.64 & 13.15 \\
        yor & 22.15 & 10.21 & 14.30 & 03.00 & 15.75 & 17.32 \\
        \hline
        \textbf{average} & 40.13 & 39.54 & 45.61 & 21.74 & 42.38 & 27.85 \\
        \hline
    \end{tabular}
    \caption{Detailed performance comparison across models on development data – Part 6.}
    \label{tab:model_comparison}
\end{table*}

\begin{table*}[h]
    \centering
    \scriptsize
    \begin{tabular}{lcccccc}
        \hline
        \textbf{language} & \textbf{XLMR-MO-LANG-FL} & \textbf{XLMR-SEP-LANG} & \textbf{XLMR-CB-ALL} & \textbf{XLMR-CB-LANG} & \textbf{XLMR-LOGREG-ALL} & \textbf{XLMR-LR-LANG} \\ \hline
        afr & 4.61 & 9.85 & 25.60 & 25.49 & 25.29 & 27.54 \\
        amh & 32.65 & 22.46 & 44.01 & 40.93 & 44.38 & 51.63 \\
        arq & 33.88 & 43.50 & 47.42 & 45.93 & 49.41 & 50.08 \\
        ary & 23.82 & 24.16 & 37.05 & 30.97 & 39.27 & 44.59 \\
        chn & 30.58 & 29.80 & 51.62 & 46.99 & 52.93 & 53.57 \\
        deu & 34.27 & 42.60 & 48.30 & 49.26 & 52.14 & 55.65 \\
        eng & 39.71 & 48.12 & 54.49 & 56.48 & 58.90 & 62.98 \\
        esp & 38.11 & 25.96 & 52.82 & 54.67 & 55.20 & 62.60 \\
        hau & 36.88 & 30.79 & 43.70 & 47.03 & 43.25 & 53.39 \\
        hin & 23.07 & 28.95 & 57.29 & 57.22 & 53.67 & 67.39 \\
        ibo & 11.86 & 23.54 & 31.22 & 27.50 & 28.34 & 37.12 \\
        kin & 23.82 & 21.50 & 28.33 & 33.82 & 26.96 & 37.34 \\
        mar & 33.57 & 28.43 & 63.88 & 67.62 & 55.17 & 73.36 \\
        orm & 24.18 & 16.65 & 31.21 & 33.78 & 34.22 & 39.94 \\
        pcm & 33.14 & 0.00 & 44.17 & 40.91 & 46.44 & 48.10 \\
        ptbr & 15.45 & 24.74 & 35.31 & 25.87 & 36.46 & 42.73 \\
        ptmz & 11.14 & 0.00 & 20.75 & 20.78 & 29.33 & 36.69 \\
        ron & 41.15 & 0.00 & 64.17 & 57.30 & 64.19 & 71.56 \\
        rus & 31.32 & 27.58 & 62.06 & 60.53 & 56.60 & 71.73 \\
        som & 22.41 & 0.00 & 31.45 & 27.94 & 36.11 & 36.56 \\
        sun & 28.34 & 28.52 & 31.23 & 29.01 & 40.34 & 37.86 \\
        swa & 17.04 & 17.71 & 19.16 & 22.58 & 23.59 & 28.14 \\
        swe & 31.95 & 43.41 & 41.73 & 41.62 & 44.29 & 48.38 \\
        tat & 23.28 & 24.31 & 35.72 & 36.54 & 36.38 & 50.66 \\
        tir & 26.44 & 25.26 & 30.24 & 30.37 & 30.94 & 36.25 \\
        ukr & 17.26 & 0.00 & 26.97 & 27.10 & 44.78 & 46.44 \\
        vmw & 11.20 & 0.00 & 5.64 & 18.07 & 10.44 & 20.13 \\
        yor & 15.81 & 7.29 & 11.78 & 18.33 & 15.51 & 23.43 \\
        \hline
        \textbf{average} & 25.61 & 21.25 & 38.48 & 38.38 & 40.52 & 46.99 \\
        \hline
    \end{tabular}
    \caption{Detailed performance comparison across models on development data – Part 7.}
    \label{tab:model_comparison}
\end{table*}

\begin{table*}[h]
    \centering
    \scriptsize
    \begin{tabular}{lcccc}
        \hline
        \textbf{language} & \textbf{XLMR-SVM-ALL} & \textbf{XLMR-SVM-LANG} & \textbf{XLMR-XGB-ALL} & \textbf{XLMR-XGB-LANG} \\ \hline
        afr & 17.13 & 29.51 & 14.79 & 18.72 \\
        amh & 24.18 & 38.36 & 38.51 & 36.58 \\
        arq & 38.47 & 35.88 & 39.46 & 35.59 \\
        ary & 24.68 & 30.92 & 26.77 & 23.71 \\
        chn & 24.88 & 40.11 & 39.64 & 35.97 \\
        deu & 38.24 & 41.24 & 42.09 & 40.91 \\
        eng & 41.83 & 46.20 & 47.89 & 43.88 \\
        esp & 35.92 & 45.50 & 53.54 & 50.61 \\
        hau & 26.81 & 39.17 & 39.11 & 39.07 \\
        hin & 27.04 & 41.26 & 62.35 & 51.74 \\
        ibo & 18.88 & 26.20 & 24.23 & 22.62 \\
        kin & 24.33 & 30.10 & 15.34 & 24.74 \\
        mar & 24.07 & 53.38 & 66.67 & 58.51 \\
        orm & 22.24 & 30.16 & 23.79 & 23.74 \\
        pcm & 35.34 & 30.80 & 32.83 & 30.10 \\
        ptbr & 23.81 & 24.92 & 22.55 & 19.10 \\
        ptmz & 14.55 & 19.99 & 6.40 & 9.26 \\
        ron & 45.39 & 55.56 & 57.30 & 50.61 \\
        rus & 33.72 & 47.71 & 54.82 & 50.04 \\
        som & 18.86 & 23.46 & 16.82 & 16.95 \\
        sun & 29.35 & 33.86 & 21.07 & 21.50 \\
        swa & 14.96 & 22.63 & 9.69 & 7.36 \\
        swe & 26.39 & 40.59 & 36.36 & 36.25 \\
        tat & 22.72 & 34.49 & 24.93 & 26.52 \\
        tir & 18.46 & 31.92 & 18.21 & 17.15 \\
        ukr & 15.68 & 21.07 & 20.91 & 14.83 \\
        vmw & 10.41 & 17.56 & 0.64 & 5.01 \\
        yor & 14.81 & 18.35 & 8.01 & 9.24 \\
        \hline
        \textbf{average} & 25.47 & 33.96 & 30.88 & 29.30 \\
        \hline
    \end{tabular}
    \caption{Detailed performance comparison across models on development data – Part 8.}
    \label{tab:model_comparison}
\end{table*}

\end{document}